\documentclass{llncs}
\usepackage{multirow}
\usepackage{amsmath}
\usepackage{amsfonts}
\usepackage{amssymb}
\usepackage{graphicx,subfig}
\DeclareMathOperator*{\argmin}{arg\,min}
\newcommand*\samethanks[1][\value{footnote}]{\footnotemark[#1]}
\newcommand{\squeezeup}{\vspace{-2.5mm}}

\begin{document}
%
\title{Deep Adversarial Context-Aware Landmark Detection for Ultrasound Imaging}


\author{Ahmet~Tuysuzoglu \inst{1}\fnmsep\thanks{Equal contribution.} \and
Jeremy~Tan\inst{2}\fnmsep\samethanks \and
Kareem~Eissa\inst{1} \and
Atilla~P.~Kiraly \inst{1}\and
Mamadou~Diallo\inst{1}\and
Ali~Kamen\inst{1}}

\institute{Siemens Healthineers, Medical Imaging Technologies, Princeton, NJ, USA\\
\and
Imperial College London}

\newcommand\blfootnote[1]{%
	\begingroup
	\renewcommand\thefootnote{}\footnote{#1}%
	\addtocounter{footnote}{-1}%
	\endgroup
}

\maketitle
\begin{abstract}

Real-time localization of prostate gland in trans-rectal ultrasound images is
a key technology that is required to automate the ultrasound guided prostate
biopsy procedures. In this paper, we propose a new deep learning based
approach which is aimed at localizing several prostate landmarks efficiently
and robustly.  We propose a multitask learning approach primarily to make
the overall algorithm more contextually aware. In this approach, we not only consider
the explicit learning of landmark locations, but also build-in a
mechanism to learn the contour of the prostate. This multitask learning is further coupled
with an adversarial arm to promote the generation of feasible structures.
We have trained this network using {\raise.17ex\hbox{$\scriptstyle\sim$}}4000 labeled
trans-rectal ultrasound images and tested on an independent set of images
with ground truth landmark locations.
We have achieved an overall Dice score of
92.6\% for the adversarially trained multitask approach, which is significantly better than the Dice
score of 88.3\% obtained by only learning of landmark locations.
The overall mean distance error using the adversarial multitask
approach has also improved by 20\% while reducing the standard deviation of the error
compared to learning landmark
locations only. In terms of computational complexity both approaches can
process the images in real-time using standard computer with a standard CUDA enabled GPU.

\end{abstract}

\blfootnote{\textbf{Disclaimer}: This feature is based on research, and is not commercially available. Due to regulatory reasons its future availability cannot be guaranteed.}

\section{Introduction}
Multi-parametric MRI can greatly improve detection of prostate cancer and can
also lead to a more accurate biopsy verdict by highlighting areas of
suspicion \cite{Boesen2017}. Unfortunately, MR-guided procedures are costly
and restrictive, whereas ultrasound guidance offers more flexibility and can
exploit the added MR information through fusion \cite{Yacoub2012}. A key step
in  the registration of diagnostic MR and live trans-rectal ultrasound is the
automatic localization of the prostate gland within the ultrasound image in
real-time. This localization could be achieved by automatically identifying a
set of image landmarks on the border of the prostate gland. This
task by itself is in general challenging due to low tissue contrast leading
to fuzzy boundaries and varying prostate gland sizes in the population.
Furthermore, prostate calcifications cause shadowing
within the ultrasound image hindering the observation of the gland
boundary. An example of this case is
shown in Fig.~\ref{fig:labelExamples}~(a).
Learning
these landmark locations is further complicated by inherent label noise as
these landmarks are not defined with absolute certainty. A small
inter-slice variability in prostate shape could result in rather larger
deviation in the landmark locations, which are placed by expert annotators.
Our analysis of this uncertainty is further explained in Section~\ref{sec:methods}.

Through initial set of experiments
we observed that individual landmark detection/regression does not yield accurate results
as the global context in terms of how the landmarks are connected is not
properly utilized. Even for expert annotators, it is
essential to use the context to place the challenging landmarks, specifically
the ones in regions with little signal or cues.
Incorporating topological/spatial priors into
landmark detection tasks is an active area of research with broad
applications. Conditional Random Fields incorporating priors
have been used with deep learning to improve delineation tasks in
computer vision \cite{Zheng15,ChenPKMY14}.
In medical imaging, improving landmark and contour localization tasks
through the use of novel deep learning
architectures has been presented in \cite{Payer16,Yangaaai17}.
In particular in \cite{Yangaaai17}, the authors considered the sequential
detection of prostate boundary through the use of recurrent neural networks
in polar coordinate transformed images; however, their method assumes that
the prostate is already localized and cropped.


In this work we propose a deep adversarial multitask
learning approach to address the challenges
associated with robust localization of prostate landmarks.
Our design aims to improve performance in regions, where the
boundary is ambiguous by using the spatial context to inform landmark placement.
Multitask learning provides an effective way to bias a
network to learn additional information that can be useful for the original
task through the use of auxiliary tasks \cite{Caruana1997}.
In particular, to bring in the global context, we learn to
predict the complete boundary contour in addition to the location of each
landmark to enforce the overall algorithm to be more
contextually aware. This multitasking network is further coupled by
discriminator network that provides feedback regarding the feasibility
of predicted contours. Our work shares similarities with
\cite{Chen2017}, where the authors used multitasking with adversarial
regularization in human pose estimation in an extensive network. Unlike
the method in \cite{Chen2017},
our approach is easily trainable and can perform at high
frame rates and compared to \cite{Yangaaai17},
it does not require the localization of the prostate gland beforehand.

\section{Methods}
\label{sec:methods}
\begin{figure*}[!tbh]
  \centering
    \subfloat[]{\raisebox{2mm}{\includegraphics[width=6.5cm]{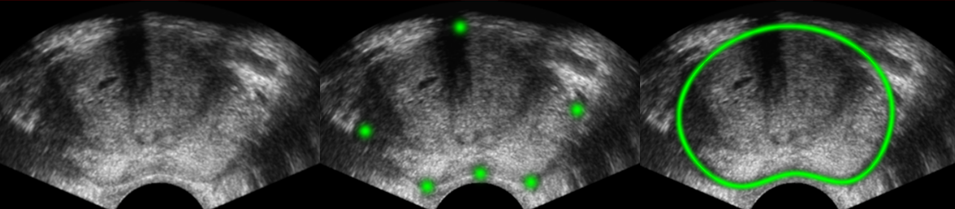}}}
    \subfloat[]{\includegraphics[width=5.5cm]{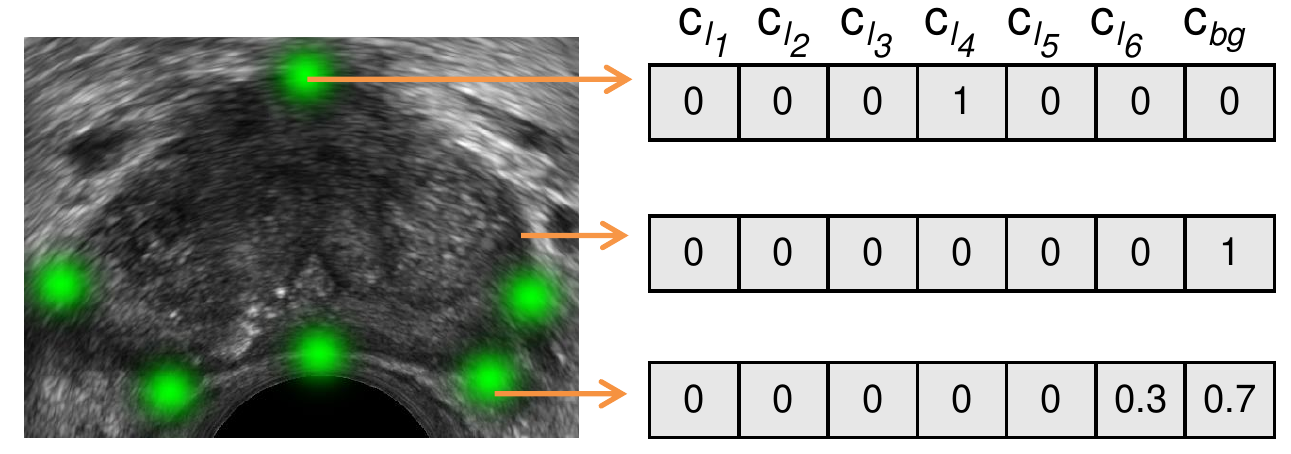}}
  \caption{(a) Ultrasound images with target labels: 2D Gaussian landmarks (center, green) and contours (right, green).\ (b) Each pixel has a distribution over 7 classes: 6 landmark classes and the background class.
Moving away from the center of a landmark, the landmark probability decreases and the background probability increases. }
~\label{fig:labelExamples}
\end{figure*}

This study includes data from trans-rectal ultrasound examinations of 32
patients, resulting in 4799 images. Six landmarks that are distributed on the
prostate boundary are marked by expert annotators. In particular, the
landmark locations are chosen to cover the anterior section of the gland (close to bladder),
posterior section (close to rectum),
and left and right extend of the gland considering the shape of the probe pressing into the
prostate. Examples of
annotations can be seen in Fig.~\ref{fig:labelExamples}~(a). Nonetheless the landmarks cannot be
placed with complete certainty due to poor boundaries, missing defining
features, shadowing and other physiological occurrences such as
calcifications. We characterized this landmark annotation uncertainty by measuring
the change in landmark position in successive frames. The mean and standard
deviation for each landmark is given in Table~\ref{errorTable}. It is understood that
part of this positional difference is due to probe and patient movement but
nevertheless they can be treated as a lower bound for the localization error
that can be achieved.

Each image is acquired as part of a 2D sweep across the prostate and all
images were resampled to have a resolution of 0.169 mm/pixel and then padded
or cropped so that the resulting image size is $512\times 512$. Training data
is tripled via augmentation with translation ($\pm$ 30-70 pixels) plus noise
($\sigma = 0.05$) and rotation ($\pm$ 4-7 $^{\circ}$) plus noise ($\sigma =
0.05$). We split the data into 3 sets: 23 patients for training (3717 images, 77\%),
6 patients for validation (853 images, 18\%), and 3 patients for testing (229 images, 5\%). For all
the methods explained below the ultrasound data is given to the network as
singe slices.
\squeezeup
\squeezeup
\subsection{Baseline Approach for Landmark Detection}
\begin{figure}
  \centering
    \includegraphics[width=\textwidth]{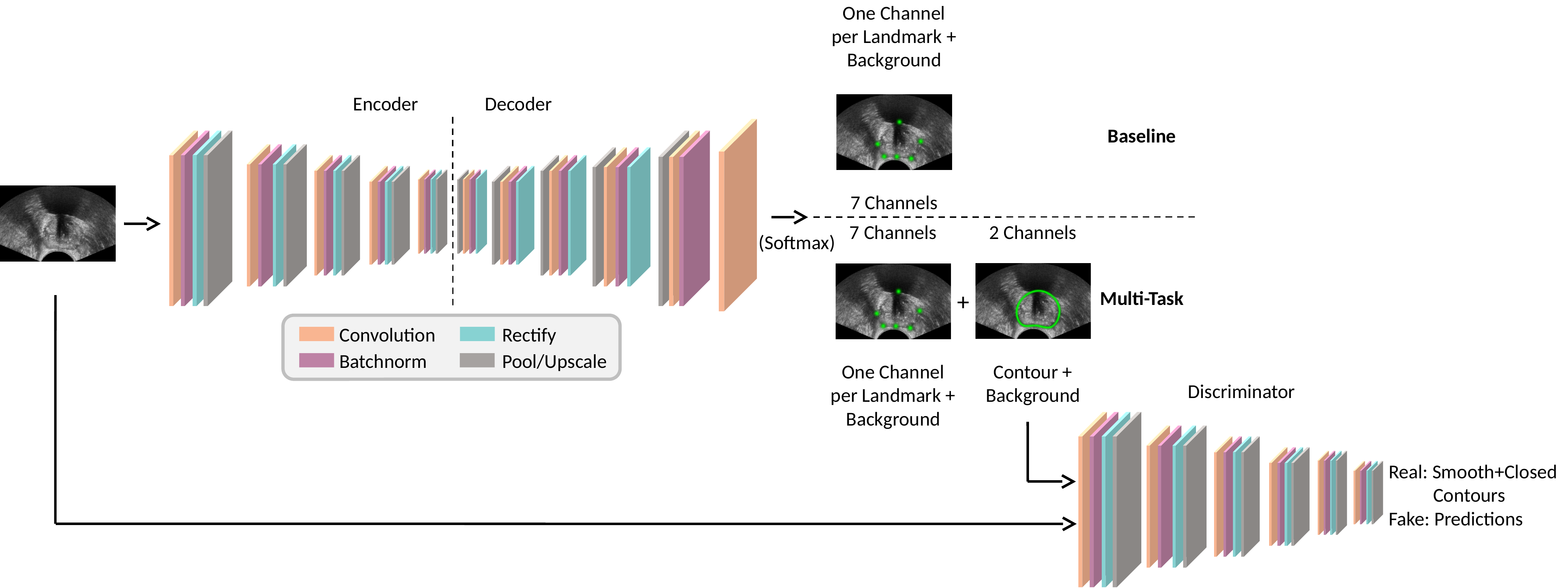}
  \caption{Our baseline network has an encoder-decoder architecture where the receptive field
  size is large enough to contain the entire prostate. The multitask network outputs
  a boundary contour along with the landmarks which is then fed to a discriminator network to evaluate
  its similarity to training set samples.}
~\label{fig:networkStructure}
\end{figure}

Given the landmark locations, our approach takes a classification approach through the
use of a shared background in
locating the landmarks rather than the classical regression approach.
The network has a 5 layer convolutional encoder and corresponding decoder with
$5\times 5$ kernels, padding of 2, stride of 1, and a pooling factor of 2 at each
layer. The number of filters in the first layer is 32; this doubles with
every convolutional layer in the encoder to a maximum of 512. The decoder
halves the number of filters with each convolutional layer. The final output
is convolved with a $1 \times 1$ kernel into 7 channels (one for each landmark and a
background class).
The configuration of the convolutional, batch normalizing,
rectifying, and pooling layers can be seen in Fig.~\ref{fig:networkStructure}.

We model each landmark as a 2D
Gaussian function centered on the landmark.
The standard deviation of this
Gaussian can in part incorporate the uncertainty involved in the landmark locations.
In contrast to the regression approaches that regress locations or
probability maps independently for each landmark, here we take a
classification approach which couples the estimation through a shared
background. For each pixel in the ultrasound image, we assign a probability
distribution over 7 classes, where we treat each landmark and the background as separate classes.
For a pixel that is at the center of a Gaussian for a landmark, the
probability for that landmark class is 1 whereas rest of the probabilities are set to zero.
These probabilities are obtained by independently normalizing each Gaussian distribution so that
the maximum of the Gaussian is 1.
Similarly for a pixel that does not overlap with any of the Gaussian functions,
the background class has probability 1 and rest of the classes are set to zero.
For a pixel that overlaps with one of the landmarks but not necessarily at the center,
the probability distribution over the classes is shared between the corresponding landmark class
and the background class. This is illustrated in Fig.~\ref{fig:labelExamples}~(b).
This framework can be trivially extended to scenarios where the Gaussian functions for the landmarks
overlap.
We learn a mapping of training images $\mathbf{x}$ in training
set $\mathbf{X}$
that represents the probability distribution of every pixel in $\mathbf{x}$ over the classes.
This mapping, $S_{\text{lm}}\left(\mathbf{x}\right)$, is learnt through the minimization
of the following supervised loss where $\mathbf{Y}_{\text{lm}}$
denotes the training set labels:


\begin{equation}
\mathcal{L}_{\text{lm}} =
-\mathbb{E}_{\left(\mathbf{x}, \mathbf{y}_{\text{lm}}\right)
\sim \left(\mathbf{X}, \mathbf{Y}_{\text{lm}}\right)}
[\log S_{\text{lm}}\left( \mathbf{x}\right)].
\label{eq:lmloss}
\end{equation}

During test time the landmark locations are obtained
by processing the output maps, i.e., by extracting the maxima.
The joint prediction of landmark and background classes could help the network become more aware of the positions of each
landmark relative to one another. However, this background class encompasses
the entire space wherever a landmark does not exist. As such, it does not
explicitly relate the points or highlight specific image features that are
relevant to the connections between points (e.g. organ contour).
\squeezeup
\squeezeup
\subsection{Multitask Learning for Joint Landmark and Contour Detection}
When deciding landmark location, expert annotators/clinicians are equipped with the prior
knowledge that the landmarks exist along the prostate boundary which is a
smooth, closed contour. Motivated by this intuition
we identify two distinct priors: First, the points lie along the prostate boundary,
and then this boundary must form a smooth, closed contour despite occlusions.
We incorporate these priors through multitask learning and the use of an adversarial
cost function.

In multitask learning, the network must identify a set of auxiliary labels
in addition to the main labels. The main labels (in this case landmarks) help
the network to learn the appearance of the landmarks; meanwhile the
auxiliary labels should promote learning of complementary cues that the
network may otherwise ignore. A fuzzy contour following the prostate prostate boundary obtained
by Gaussian blurring the spline generated by connecting the main landmark
labels is used as an auxiliary label to incorporate the first
spatial prior, that all landmarks lie on the prostate boundary.
The goal of the multitask addition is to bias the network's features
such that prostate boundary detection is enhanced. Since the boundary
overlaps directly with the landmarks, the auxiliary task lends itself well to
exploitation in the shared parameter representation. Fig.~\ref{fig:networkStructure}
displays the addition of the auxiliary label for
the multitask framework. Note that the network size does not increase,
except for the final layer, because the parameters are shared between both
tasks.

Similar to the landmark setup, we learn a mapping of training images,
$S_{\text{cnt}}\left(\mathbf{x}\right)$,
representing the likelihood of being a contour pixel
by minimizing the following supervised loss, where $\mathbf{Y}_{\text{cnt}}$
denotes the training set labels associated with the contour:
\squeezeup
\begin{equation}
\mathcal{L}_{\text{cnt}} =
-\mathbb{E}_{\left(\mathbf{x}, \mathbf{y}_{\text{cnt}}\right)
\sim \left(\mathbf{X}, \mathbf{Y}_{\text{cnt}}\right)}
[\log S_{\text{cnt}}\left( \mathbf{x}\right)].
\label{eq:cntloss}
\end{equation}
\squeezeup

\subsection*{Discriminator Network}
While the multitask framework aims to increase the network's awareness of
the prostate boundary features, it does not enforce any constraint on the
shape of the predicted contour. As such, a discriminator network is added to
motivate fulfillment of the second prior, that the boundary is a smooth
closed shape. This is helpful because the low tissue contrast can make it
challenging for the boundary detection (learned by the multitask network) to
give clean estimates without false positives. The discriminator network
is trained in a conditional style where the input training image is provided
together with the network generated or the real contour.
The design is similar to the encoder of the main encoder-decoder network with the
difference that the discriminator network is extended one layer further
and the first 3 layers have a pooling factor of 4 instead of 2. These changes
are made to rapidly discard high resolution details and focus the
discriminator's evaluation on the large scale appearance. 
We then define the discriminator loss as follows:

\squeezeup
\begin{eqnarray}
\mathcal{L}_{\text{adv}_D} =
-\mathbb{E}_{\left(\mathbf{x}, \mathbf{y}_{\text{cnt}}\right)
\sim\left(\mathbf{X}, \mathbf{Y}_{\text{cnt}}\right)}
[\log D\left(\mathbf{x},\mathbf{y}_{\text{cnt}}\right)] \nonumber \\
-\mathbb{E}_{(\mathbf{x}
\sim\mathbf{X})}
[\log \left(1- D\left(\mathbf{x}, S_{\text{cnt}}(\mathbf{x})\right)\right)].
\label{eq:disloss}
\end{eqnarray}
\squeezeup

In \cite{goodfellow2014generative}, the authors defined the
generator loss as the negative of the discriminator loss defined in
Eqn.~\ref{eq:disloss}, resulting in a min-max problem over the generator and discriminator
parameters. The authors in \cite{goodfellow2014generative} (and several others \cite{tzeng2017adversarial,usman2017stable}) have also stated the
difficulty with the min-max optimization problem and
suggested maximizing the log probability of the discriminator being mistaken as the generator loss.
This corresponds to the following adversarial loss for the landmark and contour network $S$:
\squeezeup
\begin{eqnarray}
\mathcal{L}_{\text{adv}_S} =
-\mathbb{E}_{(\mathbf{x}
\sim\mathbf{X})}
[\log D\left(\mathbf{x}, S_{\text{cnt}}(\mathbf{x})\right)].
\label{eq:advloss}
\end{eqnarray}
\squeezeup

\subsection*{Adversarial Landmark and Contour Detection Framework}
The landmark and contour detection network is trained by minimizing the following
functional with respect to its parameters $\theta_S$:
\begin{equation}
\argmin_{\theta_S}\mathcal{L}_{\text{total}} =
\mathcal{L}_{\text{lm}} + \lambda_1\mathcal{L}_{\text{cnt}}+ \lambda_2\mathcal{L}_{\text{adv}_S}
\label{eq:all}
\end{equation}
The discriminator is trained by minimizing $\mathcal{L}_{\text{adv}_D}$ with respect to its
parameters $\theta_D$.
We optimize these two losses in an alternating manner by keeping $\theta_S$ fixed in the optimization
of the discriminator and $\theta_D$ fixed in the optimization of the detector network.
In our experiments, we picked $\lambda_1=1$ and $\lambda_2=0.02$ using cross validation.

\section{Results \& Discussion}
Landmark location has a range of acceptable solutions on the prostate
boundary that is also visible in the noise of the annotated labels. As such
the Dice score between the spline interpolated prostate masks is used as the
primary evaluation metric. In addition, the Euclidean distance between
predictions and targets and the 80th percentile of this distance are
calculated. Baseline Dice score and average landmark error are 88.3\% and
3.56 mm respectively. With the multitask approach, these scores are
improved to 90.2\% and 3.12 mm respectively. The addition of adversarial
training further improves the results to 92.6\% and 2.88 mm.
In particular, note the large improvement for landmark 4 (Table
\ref{errorTable}). This is the most anterior landmark (close to bladder)
which generally has the highest error due to shadowing. Also, the improvement in the
standard deviation of the Dice score indicates that the adversarially regulated
multitask framework produces the most robust predictions.
\squeezeup
\begin{table}[h]
 \caption{Landmark annotation error together with error for baseline, multitask, and adversarial multitask methods.}
~\label{errorTable}
\begin{center}
\begin{tabular}{ |c|c|c|c|c| }
\hline
\textbf{Metric} & \textbf{Noise} & \textbf{Baseline} & \textbf{Multitask} & \textbf{Multitask GAN} \\
\hline
\hline
\multirow{6}{8em}{Mean Landmark Error $\pm$ S.D.}
&0.98 $\pm$ 0.28 & 2.11 $\pm$ 1.41 & 1.94 $\pm$ 1.36 & 1.77 $\pm$ 1.43 \\
&1.45 $\pm$ 0.44 & 2.33 $\pm$ 1.28 & 1.90 $\pm$ 1.13 & 1.97 $\pm$ 0.96 \\
&2.17 $\pm$ 0.60 & 4.03 $\pm$ 5.13 & 3.38 $\pm$ 3.68 & 3.41 $\pm$ 3.17 \\
&1.99 $\pm$ 0.47 & 6.29 $\pm$ 6.13 & 6.72 $\pm$ 5.59 & 5.01 $\pm$ 3.90 \\
&2.19 $\pm$ 0.74 & 3.44 $\pm$ 2.77 & 2.73 $\pm$ 1.94 & 3.09 $\pm$ 2.43 \\
&1.43 $\pm$ 0.54 & 3.21 $\pm$ 4.05 & 2.02 $\pm$ 1.85 & 2.01 $\pm$ 1.57 \\
\hline
\textbf{Overall Avg.} & 1.70 $\pm$ 0.51 & 3.56 $\pm$ 3.46 & 3.12 $\pm$ 2.60 & 2.88 $\pm$ 2.24 \\
\hline
\multirow{6}{8em}{80th Percentile}
& 1.42 & 3.19 & 3.04 & 2.75 \\
& 2.05 & 3.44 & 2.85 & 2.72 \\
& 3.17 & 4.59 & 4.41 & 5.08 \\
& 2.87 & 8.31 & 9.09 & 7.75 \\
& 3.14 & 4.83 & 4.27 & 4.68 \\
& 2.03 & 3.71 & 2.75 & 2.90 \\
\hline
\textbf{Overall Avg.}& 2.45 & 4.68 & 4.42 & 4.32 \\
\hline
\textbf{Avg. Dice Score $\pm$ S.D.}& - & 88.3\% $\pm$ 7.3\% & 90.2\% $\pm$ 7.2\% & 92.6\% $\pm$ 3.6\% \\
\hline
\end{tabular}
\end{center}
\end{table}

\begin{figure*}
  \centering
    \includegraphics[width=\textwidth]{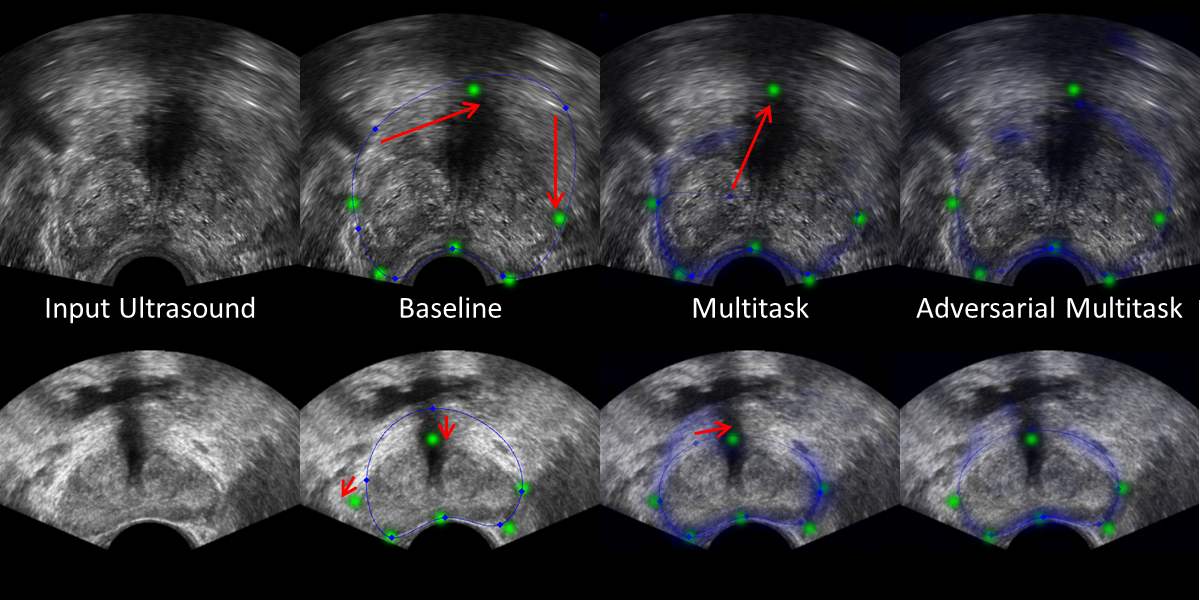}
  \caption{Ultrasound images with target (green) and prediction (blue
  diamonds, connected by spline) overlays. Red arrows indicate corrections of
  gross errors. Multitask predictions include an overlay of the contour
  prediction (blue heatmap). Adversarially regulated multitask
  learning produces more complete contours resulting in better landmark
  placement compared its plain counterpart. }
~\label{fig:USpredOverlay}
\end{figure*}

Fig.~\ref{fig:USpredOverlay} displays examples of predictions given by each method.
In the top row, the plain
multitask approach is able to improve the right-most landmark placement, but the
most anterior landmark location is still highly inaccurate.
In such cases, features learned for boundary detection can mistakenly highlight areas with high
contrast, e.g. calcification within the prostate. The adversarially trained
detector improves the landmark placement significantly. In the bottom row, the
boundary prediction is also hindered by shadowing, but the proposed framework still
improves the overall shape of the contour along with the landmark placements.

The multitask learning framework helps biasing the landmark placement toward the
prostate boundary through shared weights of two tasks namely landmark
detection and boundary estimation. As the predicted contour is not always of
high quality especially when there is signal dropouts, an adversarial
regularization is used to enhance boundary estimations and subsequently
provide a more accurate landmark detection.
\bibliography{us_dl}
\bibliographystyle{splncs03}

\end{document}